# Supervised feature evaluation by consistency analysis: application to measure sets used to characterise geographic objects

Patrick Taillandier[1,2]
[1]IRD, UMI UMMISCO 209
Bondy, France
patrick.taillandier@gmail.com

Alexis Drogoul[1,2,3]
[2]IFI, MSI, UMI 209,
Ha Noi, Viet Nam
alexis.drogoul@gmail.com

[3]UPMC, UMI 209,
Paris, France

*Abstract*—Nowadays, supervised learning is commonly used in many domains. Indeed, many works propose to learn new knowledge from examples that translate the expected behaviour of the considered system. A key issue of supervised learning concerns the description language used to represent the examples. In this paper, we propose a method to evaluate the feature set used to describe them. Our method is based on the computation of the consistency of the example base. We carried out a case study in the domain of geomatic in order to evaluate the sets of measures used to characterise geographic objects. The case study shows that our method allows to give relevant evaluations of measure sets.

***Supervised feature evaluation; consistency computation; geomatic***

## I. Introduction (Heading 1)

Supervised learning is more and more used in numerous domains. Most of these works use an "attribute-value" formalism to represent examples. Thus, examples are represented by a set of feature values. The choice of this feature set is a key issue in supervised learning. Indeed, if the notion that we search to learn is not "contained" in the feature set, it will be not possible to learn it. For example, if we search to learn the concept of "square object" but if the feature set used to describe the objects contains no information about the shape of the objects, but only information concerning their colour, it is then not possible to learn this concept. Higher the number of features, more the learning is complex, and more examples are needed to learn. Thus, it is important to have enough feature to well represent the examples, but at the same time, to limit the number of features to ease the learning. In this context, be able to evaluate feature sets is particularly interesting. In this paper, we propose a supervised method to evaluate a feature set. Our method is based on the computation of the example base consistency.

In Section II, we describe the context of this work. Section III is dedicated to the presentation of our feature set evaluation method. Section IV presents a case-study carried out in the domain of geomatic. At last, Section V concludes and gives perspectives.

## II. Context

### A. Formalisation of the problem

We are interested in the evaluation of a feature set considering a given task. Indeed, a feature set can be relevant for a given task, but not at all for another one.

In this work, we assume that an example base describing the expected behaviour of the considered system is defined. Examples are described by a vector of feature values and by a label. The vector of feature values describes the example characteristics, and the label, the concept that we search to learn. In this work, we assume that the set of possible labels is finite. Figure 1 gives an example of an example base that concerns the learning of procedural knowledge for a geomatic process: the goal is to learn knowledge relative to the choice of the best operation to apply on building groups according to their geometric characteristics to produce a 50k map. The goal of the supervised feature set evaluation is then to compute, from the example base, a quality value for the feature set.

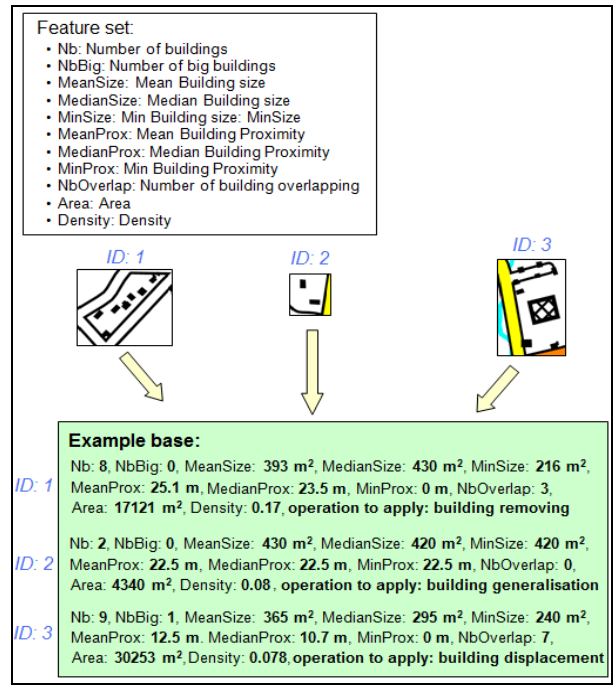

Figure 1. Example of an example base for building group

### B. Evaluation of feature sets

In the literature, many methods were proposed to evaluate a set of features from an example base. These methods can be divided in two major groups: the ones that work on individual features, and the ones that take into account all the features at the same time. Concerning the first groups of methods, many of them are based on classic measures such as the chi-squared statistic or the entropy gain. Other methods such as [1] propose to analysis the linked between the variations of a feature value and the label associated. For the second group of method, a classic strategy consists in evaluating the predictive model built from the example base (often by cross-validation [2]). Other methods imply to compute the degree of redundancy between features [3]. Most of these methods search to estimate the worth of subsets of features in order to be able to remove useless features, and thus to ease the learning. In this paper, we propose a method that will focus on evaluating the necessity to add new features. Indeed, our goal will be more to determine if more features are needed rather than removing useless ones.

## III. METHOD PROPOSED

The method we propose for the supervised evaluation of feature sets is based on the computation of the inconsistency of the example base. Inconsistency means, here, cases where, for a same state (according to the feature set), different labels are assigned to the examples. For example, if an example base contains two examples representing two building groups that are similar in terms of feature values, and if the operation that has to be applied on the first building group is a *building removing operation*, and on the second one, a *building displacement operation*, the example base is inconsistent.

Our feature set evaluation method is based on three steps:
- Selection of the pertinent features.
- Discretisation of the selected features.
- Computation of the inconsistency from the resulting example base.

### A. Selection of the pertinent features

As states in Section II.B, the goal of our method is to determine is a feature set characterises well enough the examples or if more features are needed.

In order to achieve this goal, we propose as a first step, to remove irrelevant features that only bring noise to the learning process. Indeed, some features can be totally unrelated to the concept we try to learn.

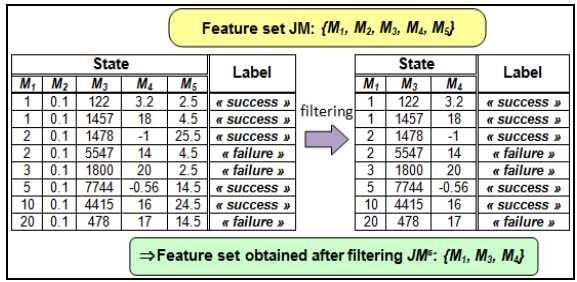

Figure 2. Example of pertinent feature selection

For example, colour information is useless to learn the concept of square object. In the literature, many supervised filtering algorithms were proposed (see [4]). These methods allow to select a subset of pertinent features from a example base. Most of them are based on the measures presented in Section II.B. The application of such algorithms on the example base allows to obtain a new feature set composed of features that are relevant to the concept we try to learn. Figure 2 gives an example of feature set filtering.

### B. Discretisation of the selected features

As a second step, we propose to discretise the numeric features. Similarly to the feature filtering problem, numerous works deals with the problem of supervised feature discretisation (e.g. [5] or [6]). These algorithms propose, by example base analysis, to pass from continuous feature to nominal ones composed of sets of disjoint intervals. The goal is to decrease the size of the feature set space and to make finite the number of possible example descriptions. Indeed, some features can allow slight variation of their values that does not justify a different treatment of the examples. Figure 3 gives an example of feature set discretisation.

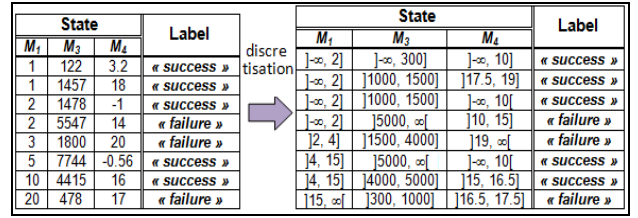

Figure 3. Example of feature set discretisation

### C. Computation of the inconsistency from the resulting example base

At the end of the two previous steps, we obtain a new example base that we use to evaluate the feature set. We propose for that to evaluate the inconsistency rate of the resulting example base. An inconsistent example base is a base in which examples, with the same feature values, have different labels. An example base strongly inconsistent means that additional features are needed to characterise the examples.

Let $S_B$ be the set of different feature descriptions (according to the feature values) contains in the examples base $B$. Thus, $S_B$ contains a unique specimen, in terms of feature values, of each example contained in $B$. We define the function $nb\_examples_B(s,l)$ that returns the number of examples in the example base $B$ that have $s$ for feature description and $l$ for label. The inconsistency rate of an example base $B$ for which the set of possible labels is $L$ is computed by the following formulae:

$$\text{Inconsistency}(B) = 1 - \frac{\sum_{s \in S_B} \left[ \max_{l \in L}(nb\_examples_B(s,l)) \right]}{card(B)}$$

Figure 4 gives an example of inconsistency computation for two examples bases.

The inconsistency rate is bounded by:

$$0 \leq \text{Inconsistency}(B) \leq 1 - \frac{1}{card(L)}$$

Indeed, at best, for each different feature description of the example base, only one label is proposed. Thus, we have:

$$\sum_{s \in S_B} \left[ \max_{l \in L}(nb\_examples_B(l,s)) \right] = card(B)$$

The inconsistency rate is thus equals to 0, which means that the example base is totally consistent.

At worst, for each different feature descriptions of the example base, each possible label is represented with the same proportion in $B$. Thus, for each different feature description $s$, we can compute:

$$\max_{l \in L}(nb\_examples_B(s,l)) = \frac{1}{card(L)}$$

Thus, the inconsistency rate is equals to *(1 – 1/card(L))*, which means that the example base is totally inconsistent.

Higher the inconsistency rate of an example base, more this one suffers from a description noise, and thus, more it is important to add new relevant features to characterise the examples.

| $B_{ex1}$ | State | | | Label | $B_{ex2}$ | State | | | Label |
|---|---|---|---|---|---|---|---|---|---|
| | $M_1$ | $M_3$ | $M_4$ | | | $M_1$ | $M_3$ | $M_4$ | |
| | ]-∞, 2] | ]-∞, 300] | ]-∞, 10] | « success » | | ]-∞, 2] | ]1000, 1500] | ]17.5, 19] | « success » |
| | ]-∞, 2] | ]1000, 1500] | ]17.5, 19] | « success » | | ]-∞, 2] | ]1000, 1500] | ]17.5, 19] | « success » |
| | ]-∞, 2] | ]1000, 1500] | ]-∞, 10] | « failure » | | ]-∞, 2] | ]1000, 1500] | ]17.5, 19] | « failure » |
| | ]-∞, 2] | ]5000, ∞[ | ]10, 15] | « failure » | | ]-∞, 2] | ]5000, ∞[ | ]10, 15] | « failure » |
| | ]2, 4] | ]1500, 4000] | ]19, ∞[ | « failure » | | ]-∞, 2] | ]5000, ∞[ | ]10, 15] | « success » |
| | ]4, 15] | ]5000, ∞[ | ]-∞, 10] | « success » | | ]4, 15] | ]5000, ∞[ | ]-∞, 10] | « success » |
| | ]4, 15] | ]4000, 5000] | ]15, 16.5] | « success » | | ]4, 15] | ]4000, 5000] | ]15, 16.5] | « success » |
| | ]15, ∞[ | ]300, 1000] | ]16.5, 17.5] | « failure » | | ]15, ∞[ | ]300, 1000] | ]16.5, 17.5] | « failure » |

$S_{Bex1} = \{S_1=(M_1\in]-\infty, 2] \land M_3\in]-\infty, 300] \land M_4\in]-\infty, 10]);$
$S_2=(M_1\in]-\infty, 2] \land M_3\in]1000, 1500] \land M_4\in]17.5, 19]);$
$S_3=(M_1\in]-\infty, 2] \land M_3\in]1000, 1500] \land M_4\in]-\infty, 10]);$
$S_4=(M_1\in]-\infty, 2] \land M_3\in]5000, \infty[ \land M_4\in]10, 15]);$
$S_5=(M_1\in]2, 4] \land M_3\in]1500, 4000] \land M_4\in]19, \infty[);$
$S_6=(M_1\in]4, 15] \land M_3\in]5000, \infty[ \land M_4\in]-\infty, 10]);$
$S_7=(M_1\in]4, 15] \land M_3\in]4000, 5000] \land M_4\in]15, 16.5]);$
$S_8=(M_1\in]15, \infty[ \land M_3\in]300, 1000] \land M_4\in]16.5, 17.5])\}$

$Inconsistency(B_{ex1}) = 1 - \frac{\max[nb\_examples_B(S_1, success); nb\_examples_B(S_1, failure)] + ...}{card(B_{ex1})}$

$Inconsistency(B_{ex1}) = 1 - \frac{\max(1;0) + \max(1;0) + \max(1;0) + \max(0;1) + \max(0;1) + \max(1;0) + \max(1;0) + \max(0;1)}{8}$

$Inconsistency(B_{ex1}) = 1 - \frac{1+1+1+1+1+1+1+1}{8} = 0$

$S_{Bex2} = \{S_2=(M_1\in]-\infty, 2] \land M_3\in]1000, 1500] \land M_4\in]17.5, 19]);$
$S_4=(M_1\in]-\infty, 2] \land M_3\in]5000, \infty[ \land M_4\in]10, 15]);$
$S_6=(M_1\in]4, 15] \land M_3\in]5000, \infty[ \land M_4\in]-\infty, 10]);$
$S_7=(M_1\in]4, 15] \land M_3\in]4000, 5000] \land M_4\in]15, 16.5]);$
$S_8=(M_1\in]15, \infty[ \land M_3\in]300, 1000] \land M_4\in]16.5, 17.5])\}$

$Inconsistency(B_{ex2}) = 1 - \frac{\max[nb\_examples_B(S_2, success); nb\_examples_B(S_2, failure)] + ...}{card(B_{ex2})}$

$Inconsistency(B_{ex2}) = 1 - \frac{\max(2;1) + \max(1;1) + \max(1;0) + \max(1;0) + \max(0;1)}{8}$

$Inconsistency(B_{ex2}) = 1 - \frac{2+1+1+1+1}{8} = 0.25$

Figure 4.   Example of inconsistency computation

## IV. 4. CASE STUDY

### A. General case-study context

#### 1) Supervised learning in geomatic

The introduction of Artificial Intelligence techniques in cartography has allowed to automate many processes. In particular, Artificial Intelligence has brought new techniques to acquire knowledge from examples. Supervised learning techniques are now widely used in Geomatics: for acquisition of procedural knowledge (e.g. [7] or [8]), for data enrichment (e.g. [9] or [10]), system calibration (e.g. [11]), etc. Most of these works use an "attribute-value" formalism to represent examples. Thus, examples are represented by a set of measures characterising the geographic object states.

#### 2) Automated cartographic generalisation

In order to evaluate our method, we present a case study we carried out in the domain of cartographic generalisation. Cartographic generalisation is the process that aims at simplifying geographic data to suit the scale and purpose of a map. Its automation is a complex problem. One approach to solve it consists in using a local, step-by-step and knowledge-based method ([7] and [12]). A classic generalisation model based on such approach is the AGENT model ([13] and [14]). It proposes to model geographic objects (or groups of geographic objects) by agents, i.e. autonomous entities that manage their own generalisation. In order to generalise itself, the geographic agent chooses actions (i.e. generalisation operations) to apply according to procedural knowledge and to its state. The agent states are characterised by a set of measures that depends of the nature of the considered knowledge. The agents can also evaluate their satisfaction, which characterises the respect of cartographic constraints (map specifications) by its state. For example, a cartographic constraint can be for a building agent to be big enough to be readable. Each constraint has a satisfaction level that is ranged between 1 and 10 (10 means that the constraint is perfectly satisfied). The agent satisfaction consists in a linear combination of its constraint satisfaction weighted by their importance.

#### 3) Procedural knowledge revision

In [15], we proposed an approach to revise the procedural knowledge of systems based on the AGENT model. This approach allows to build example bases (one per element of knowledge) from experience. These example bases are then used to search the knowledge that maximises the performance of the generalisation system.

Three types of knowledge are concerned by the knowledge revision:

- *Knowledge relative to action application*: for each action, a set of rules (that depends of a set of measures), defines if the action has to be applied or not, and with which priority.
- *Stopping criterion*: a set of rules defines, according to a given state and to the initial state, if the agent has to try to apply more actions or not. If no specific stopping criterion is defined, the generalisation process ends when all actions have been tested for all states or when a perfect state (i.e. satisfaction = 10) has been found.
- *Optimality criterion*: a set of rules defines, according to a given state, if this state can be improved or not, i.e. if the application of other actions can allow to obtain a better state or not. This criterion is different to the stopping criterion because it allows to backtrack to previous states and to test other actions. This criterion marks a local optimal while the stopping criterion marks a global optimal.

### B. Revision of knowledge used by building group agent

#### 1) Building group generalisation

The case study we propose focuses on the generalisation of building groups: a building group is a space composed of a set of "close" buildings belonging to the same building block (space surrounded by a minimum cycle of roads). The initial data are stemming from BD TOPO®, the 1m resolution database of the French NMA (reference scale approximately 1:15 000). The target scale is 1:50 000.

We defined six constraints for the building group:
- *Proximity constraint*: states that the buildings should be neither too close to each other, nor too close to the roads.
- *Building satisfaction constraint*: states that the buildings composing the building group should individually satisfy their internal constraints.
- *Density constraint*: states that the building density is not too high in comparison to the initial building density.

- **Spatial distribution constraint**: states that the current building spatial distribution should be close to the initial building spatial distribution.
- **Big building preservation constraint** states that the buildings, of which the area is bigger than a threshold, should not be removed.
- **Corner building preservation constraint**: states that the buildings located at the corner of roads should not be removed.

In order to satisfy its constraints, the *building group* agent can apply five actions:
- **Building generalisation action**: triggers the individual generalisation of the building agents composing the building group.
- **Building displacement action**: displaces buildings that have proximity problems.
- **Local building removal action**: removes buildings according to a local context.
- **Building removal/displacement action**: selects the building that has the most serious proximity problems and removes it.
- **Global building removal action**: removes buildings according to the global context.

TABLE I.  INITIAL KNOWLEDGE

|  | Knowledge |
|---|---|
| **Stopping criterion** | Never stop the process |
| **Optimality criterion** | A state is never optimal |
| **Building generalisation action** | Always apply this action |
| **Building displacement action** | Always apply this action |
| **Local building removal action** | Always apply this action |
| **Building removal/displacement action** | Always apply this action |
| **Global building removal action** | Always apply this action |

The initial knowledge used for the experiment is presented in Table I. It tests all the possible actions for each state and does not define optimality nor stopping criterion. Thus, it ensures to find for each generalisation the best possible state considering the available constraints and actions. Nevertheless, it requires exploring many states per generalisation and is thus not efficient at all.

*2) Feature sets defined*

We defined 22 measures for the building group agent:
- *random*: return a random number ranged between 0 and 1
- *id*: building group identifier in the database
- *sat_proxy*: satisfaction of the proximity constraint
- *sat_building_sat*: satisfaction of the building satisfaction constraint
- *sat_density*: satisfaction of the density constraint
- *sat_spatial*: satisfaction of the spatial distribution constraint
- *sat_large*: satisfaction of the large building preservation constraint
- *sat_corner*: satisfaction of the corner building preservation constraint
- *nb_bldg*: number of buildings
- *nb_large_bldg*: Number of large buildings (larger than a threshold)
- *nb_corner_bldg*: Number of buildings located in the corner of roads
- *area*: Building group area
- *density*: Building density
- *min_sat*: Minimum of the individual building satisfaction
- *median_sat*: Median of the individual building satisfaction
- *first_quartile_sat*: First quartile value of the individual building satisfaction
- *min_dist*: Minimum distance between buildings
- *median_dist*: Median distance between buildings
- *first_quartile_dist*: First quartile distance value between buildings
- *mean_dist*: Mean distance between buildings
- *nb_overlaps*: Number of buildings overlapping other buildings or roads
- *ratio_overlaps*: Ratio of building surface overlapping other buildings or roads.

In order to test our evaluation method, we defined three groups of measure (feature) sets:
- *Irrelevant measure sets (I)*: composed of measures that give no information about the knowledge. The interest to test these measures set is to check that our evaluation method allow to detect irrelevant measure sets.
- *Basic measure sets (B)*: composed of the constraint satisfaction. The interest to test these measure sets comes from the fact that these measures are often used to defined procedural knowledge (e.g. [16] and [17])
- *Complete measure set (C)*: composed of the constraint satisfaction and of additional measures.

Table II gives the detailed composition of each measure set.

TABLE II.  MEASURE SETS DEFINED

|  | I | B | C |
|---|---|---|---|
| **Stopping criterion** | •random<br>•id | •sat_proxy<br>•sat_building_sat<br>•sat_density<br>•sat_spatial<br>•sat_large<br>•sat_corner | All measures except:<br>•random<br>•id |
| **Optimality criterion** | •random<br>•id | •sat_proxy<br>•sat_building_sat<br>•sat_density<br>•sat_spatial<br>•sat_large<br>•sat_corner | All measures except:<br>•random<br>•id |
| **Building generalisation action** | •random<br>•id | •sat_building_sat | •sat_building_sat<br>•min_sat<br>•median_sat<br>•first_quartile_sat |
| **Building displacement action** | •random<br>•id | •sat_building_sat | •sat_proxy<br>•min_dist<br>•median_dist<br>•first_quartile_dist<br>•mean_dist<br>•nb_overlaps<br>•ratio_overlaps |
| **Local building removal action** | •random<br>•id | •sat_building_sat | •sat_proxy<br>•min_dist<br>•median_dist<br>•first_quartile_dist |

| | | | •mean_dist<br>•nb_overlaps<br>•ratio_overlaps |
|---|---|---|---|
| **Building removal/displacement action** | •random<br>•id | •sat_building_sat | •sat_proxy<br>•min_dist<br>•median_dist<br>•first_quartile_dist<br>•mean_dist<br>•nb_overlaps<br>•ratio_overlaps |
| **Global building removal action** | •random<br>•id | •sat_density | •sat_density<br>•nb_bldg<br>•area<br>•density |

*3) Results*

Table III presents the results of the *irrelevant measure sets*, Table IV of the *basic measure sets* and Table V of the *complete measure sets*. For each group of measure sets, we present for each element of knowledge:

- *Measures*: The measures used by the rules obtained after revision.
- *Performance*: The performance of the generalisation system while using the revised knowledge. This score was computed in a different area than the one used to revise the knowledge (and to compute the inconsistency rate of the measure sets). The test area is composed of 200 building groups and the revision was carried out from the generalisation of 50 building groups. The function uses to compute the system performance favours the effectiveness of the system (the cartographic quality of the results) over its efficiency (speed to carry out the generalisation process). Concerning the action application knowledge, only one score is presented for all actions because we revised all these elements of knowledge at the same time.
- *Minority*: The proportion of examples belonging to the minority label. Concerning the elements of knowledge relative to action application, two labels are defined: "apply this action" and "do not apply this action". For the stopping criterions, the two labels are: "stop the generalisation process" and "continue the generalisation process". At last, for the optimality criterion, the two labels are: "optimal state" and "non-optimal state". The interest of this minority label proportion value comes from the fact that it gives an indication for the maximum value of the inconsistency rate. Indeed, in the case where no measure is relevant, no measure is kept after the measure selection step. Thus, card($S_B$)=1, and as only two different labels are defined for our elements of knowledge, the inconsistency rate is then bounded by:

$$Inconsistency(B) \leq 1 - \frac{\max[nb\_examples_B(s,label_1), nb\_examples_B(s,label_2)]}{nb\_examples_B(s,label_1) + nb\_examples_B(s,label_2)}$$

$$\Leftrightarrow Inconsistency(B) \leq \frac{\min[nb\_examples_B(s,label_1), nb\_examples_B(s,label_2)]}{nb\_examples_B(s,label_1) + nb\_examples_B(s,label_2)}$$

- *Inconsistency*: The inconsistency rate computed by our method

For the *irrelevant measure sets* (Table III), the inconsistency rate is always equal to the proportion of the minority label, which shows that the measures bring no information about the knowledge. This statement is totally consistent with the definition of these measure sets. We can note as well that no measure is used by the revised rules. At last, we can note that the performance after revision of the knowledge relative to the action application has been improved while no measure was used. In fact, the rule "always apply the *building removal/displacement* action" was replaced by the rule "never apply the *building removal/displacement* action", and the rule "always apply the *global building removal* action" was replaced by the rule "never apply the *global building removal* action". The replacement of these rules allowed to improve the efficiency of the system and thus its performance.

TABLE III. RESULTS FOR THE *IRRELEVANT MEASURE SETS*

| | Measures | Performance | Minority | Inconsistency |
|---|---|---|---|---|
| **Stopping criterion** | Not any | 0.761 | 0.29 | 0.29 |
| **Optimality criterion** | Not any | 0.761 | 0.32 | 0.32 |
| **Building generalisation action** | Not any | 0.784 | 0.12 | 0.12 |
| **Building displacement action** | Not any | | 0.41 | 0.41 |
| **Local building removal action** | Not any | | 0.16 | 0.16 |
| **Building removal/displacement action** | Not any | | 0.04 | 0.04 |
| **Global building removal action** | Not any | | 0.3 | 0.3 |

TABLE IV. RESULTS FOR THE *BASIC MEASURE SETS*

| | Measures | Performance | Minority | Inconsistency |
|---|---|---|---|---|
| **Stopping criterion** | Not any | 0.761 | 0.29 | 0.19 |
| **Optimality criterion** | • sat_corner | 0.763 | 0.32 | 0.14 |
| **Building generalisation action** | Not any | 0.784 | 0.12 | 0.12 |
| **Building displacement action** | Not any | | 0.41 | 0.36 |
| **Local building removal action** | Not any | | 0.16 | 0.15 |
| **Building removal/displacement action** | Not any | | 0.04 | 0.04 |
| **Global building removal action** | Not any | | 0.3 | 0.3 |

For the *basic measure sets* (Table IV), the inconsistency rates are still very close to the proportion of the minority label. The constraint satisfaction brings little information about the knowledge. The only elements of knowledge for which the inconsistency rate has significantly decreased are the stopping and the optimality criteria. It is for the optimality criterion that this rate has the most decreased and it is the only element of knowledge that uses measures and for which the performance has improved in comparison to the previous measure sets. Thus, we can state a correlation between the decrease of the inconsistency rate and the fact that a relevant measure has been added to the measure set. At last, we can deduce from this second experiment that the use of measure sets uniquely composed of constraint satisfaction can be insufficient for knowledge definition.

TABLE V. RESULTS FOR THE *COMPLETE MEASURE SETS*

| | Measures | Performance | Minority | Inconsistency |
|---|---|---|---|---|
| **Stopping criterion** | •*sat_spatial*<br>•*sat_proximity*<br>•*min_sat* | 0.768 | 0.29 | 0.14 |
| **Optimality criterion** | •*sat_corner* | 0.763 | 0.32 | 0.14 |
| **Building generalisation action** | Not any | 0.786 | 0.12 | 0.12 |
| **Building displacement action** | •*min_dist* | | 0.41 | 0.32 |
| **Local building removal action** | Not any | | 0.16 | 0.15 |
| **Building removal/displacement action** | Not any | | 0.04 | 0.04 |
| **Global building removal action** | •density<br>•nb_bldg | | 0.3 | 0.15 |

For the *complete measure sets* (Table V), the adding of pertinent measures has allowed to decrease the inconsistency rate for some of the elements of knowledge. Concerning the elements of knowledge relative to the application of the *Local building removal* action and to the *Building removal/displacement* action, the rate remained (almost) equal to the proportion of examples of the minority label. In fact, the new measures did not bring much information about these elements of knowledge. For the other elements of knowledge, the inconsistency rate has decreased and rules using measures has been defined. We can note that these rules are pertinent because they allowed an improvement of the system performance. To conclude on this last group of measure sets, we can note that they are pertinent, but they can be improved. Typically, it could be interesting to define new measures for the element of knowledge relative to the *building displacement* action for which the inconsistency rate is still high.

These experiments allowed to show the relevance of our measure set evaluation method. Indeed, it allowed to diagnose which measure sets were pertinent for the definition of good knowledge. These experiments showed as well that the use of measure sets uniquely composed of the constraint satisfaction can be insufficient to define good knowledge. The adding of supplementary measures is often necessary. At last, these experiments allowed to point out the measure sets that needs more urgently to be improved.

## V. CONCLUSION

In this paper, we proposed a method dedicated to the supervised evaluation of feature sets. This method allows to give a diagnostic of feature sets by analysing the inconsistency of example bases. We showed the pertinence of our method through a case-study carried out in the domain of geomatic.

As our method is generic and can be used as soon as an example base is built, an interesting perspective could consist in testing it for other applications domain than geomatic.

Another interesting perspective could be to define, from this feature set evaluation method, a complete methodology for feature set definition. Indeed, by detecting the need of new features, our method could be part of a broader feature set definition methodology that could help experts to define their own feature set.